# Egalitarian Language Representation in Language Models: It All Begins with Tokenizers

**Menan Velayuthan and Kengatharaiyer Sarveswaran**

Department of Computer Science, University of Jaffna, Sri Lanka

vmenan95@gmail.com, sarves@univ.jfn.ac.lk

## Abstract

Tokenizers act as a bridge between human language and the latent space of language models, influencing how language is represented in these models. Due to the immense popularity of English-Centric Large Language Models (LLMs), efforts are being made to adapt them for other languages. However, we demonstrate that, from a tokenization standpoint, not all tokenizers offer fair representation for complex script languages such as Tamil, Sinhala, and Hindi, primarily due to the choice of pre-tokenization methods. We go further to show that pre-tokenization plays a more critical role than the tokenization algorithm itself in achieving an egalitarian representation of these complex script languages. To address this, we introduce an improvement to the Byte Pair Encoding (BPE) algorithm by incorporating graphemes, which we term Grapheme Pair Encoding (GPE). Our experiments show that grapheme-based character extraction outperforms byte-level tokenizers for complex scripts. We validate this approach through experiments on Tamil, Sinhala, and Hindi.

## 1 Introduction

Large Language Models (LLMs) have gained significant attention from the research community and the general public, especially following the public release of OpenAI's ChatGPT in 2022 (OpenAI, 2022). LLMs have been heralded for their economic impact (Teubner et al., 2023; Eloundou et al., 2023) and have applications in areas such as coding assistants (Rozière et al., 2024; Li et al., 2023), chat systems (Meyer et al., 2023; Dortheimer et al., 2024), and machine translation (Brants et al., 2007; Dabre et al., 2020).

Mainstream LLMs are pretrained on English-dominant corpora(Zhao et al., 2024), which makes them English-centric (EC) models. Among these, Llama 3 (Dubey et al., 2024), Mistral (Jiang et al., 2023), and Phi-3 (Abdin et al., 2024) are popular in the research community due to their open weights and strong performance on tasks like the Massive Multitask Language Understanding (MMLU) (Hendrycks et al., 2021). This popularity has led to the use of EC LLMs as base models for developing non-English language models, such as Tamil Llama (Balachandran, 2023) and Llama 3 8B CPT SEA-LIONv2 (Lowphansirikul et al., 2021; Singapore, 2024). However, their limitations in representing languages with complex scripts may be overlooked due to their widespread use.

Petrov et al. (2024) demonstrates that the unequal treatment of languages begins at the tokenizer level. They argue that commercial LLM services charge based on token count, meaning that languages requiring more tokens may disadvantage users interacting in those languages. They also discuss how tokenization differences impact latency and context window requirements, as a higher token count necessitates more computational time and larger context windows. Our observations reveal that pre-tokenization is a key factor contributing to the unfair representation of languages with complex scripts such as Tamil, Sinhala, and Hindi. We argue that pre-tokenization limits the number of tokens a tokenizer can learn (refer to Section 3.1). By analyzing pre-tokenization, we can simulate the effect of a tokenizer trained with sufficient data on these languages. This helps assess whether using such a tokenizer for adapting to new complex script languages is equitable. Additionally, we show that pre-tokenization plays a more significant role than the choice of the tokenization algorithm itself (refer to Figure 1).

Byte Pair Encoding (BPE) (Sennrich et al., 2016) is a popular choice for tokenizers in English-centric LLMs. However, applying BPE directly to languages with complex scripts may not be optimal. To address this, we modified the BPE algorithm by using graphemes (refer to Section 3.2) as the atomic units, which we call Grapheme Pair Encoding (GPE). This adaptation enables BPE to recognize characters in complex scripts similar to how humans would realize them, resulting in improved performance compared to standard BPE.

Efforts have been made to develop tokenizer-free language models (Deiseroth et al., 2024; Yu et al., 2023), which operate at the byte level. However, this approach may not be ideal for languages with complex scripts such as Tamil, Sinhala, and Hindi. Processing these languages using graphemes proves to be more beneficial. We compare byte-level tokenizers, such as ByT5 (Xue et al., 2022) and CANINE (Clark et al., 2022), with grapheme-based character extractors and demonstrate that the latter performs better on our evaluation metrics.

## 2 Related Works

**Evaluating Tokenizers.** Goldman et al. (2024) examine the correlation between tokenizers' text compression and language models' performance on downstream tasks, showing a significant relationship. They conclude that compression is a reliable intrinsic indicator of tokenization quality. Rust et al. (2021) propose two metrics: 1) Normalized Sequence Length (NSL), which compares a tokenizer's compression against a baseline like Llama, and 2) Bytes per Token, calculated by dividing UTF-8 bytes by the tokens produced. While NSL is relative, we use a variant of Bytes per Token as an absolute measure. Petrov et al. (2024) introduce *Tokenization Parity* ($TP$), a metric assessing the tokenization of one language relative to another. We use $TP$ to evaluate how complex scripts are underrepresented compared to English. Rust et al. (2021) also propose "subword fertility" and "proportion of continued words" as additional measures. Given their correlation with the previously discussed metrics, we choose "Compression Ratio" and "Tokenization Parity" for our evaluation.

**Tokenization Algorithms.** Sennrich et al. (2016) introduce Byte Pair Encoding (BPE) as a subword segmentation strategy to manage open-vocabulary challenges in neural machine translation. This method, which improves translation accuracy by breaking down rare and unknown words into sequences of subword units, is derived from a data compression algorithm. It segments words into the most frequent pairs of bytes, facilitating a compact representation of open vocabularies using a fixed-size vocabulary of subword units. BPE-based tokenizers are a popular choice for English-centric Large Language Models (LLMs). However, they may perform suboptimally with complex scripted languages that require combinations of more than two Unicode codepoints to create characters, especially when trained with limited data and a fixed vocabulary size. Other commonly used subword tokenization techniques include WordPiece (Wu et al., 2016) and the Unigram algorithm (Kudo, 2018).

## 3 Background

This section introduces pre-tokenization and graphemes to support understanding of the following content.

### 3.1 Pre-tokenization

Although tokenizers can be trained on large strings of textual data using Byte Pair Encoding (BPE), training them naively may lead to suboptimal performance on certain downstream tasks as these tokenizers may lead to tokens forming around common phrases or sentence (Dagan et al., 2024). To address this, it is beneficial to include a pre-tokenization step, a preprocessing step before the actual tokenization, that breaks the input text into smaller, manageable chunks. We will refer to these as *pre-tokens* throughout the paper. Since it's inception, numerous pre-tokenization methods have been proposed, such as splitting on punctuation marks or spaces, using linguistic rule-based approaches, and regular expression-based methods (Dagan et al., 2024). HuggingFace (HF) supports various pre-tokenization techniques, and further details can be found on the `pretokenization` page[1]. Table 1 illustrates the pre-tokenization outputs for the transla-tion of the text "hello!" in English, Tamil, Sinhala, and Hindi. It is evident that models like GPT-2 (Radford et al., 2019), GPT-4 (OpenAI et al., 2023), and Llama 3 (Dubey et al., 2024) unnecessarily break the text for Tamil, Sinhala, and Hindi, resulting in the need for a larger context window.

---

[1] https://huggingface.co/docs/tokenizers/en/api/pre-tokenizers

| Tokenizer | en | ta | si | hi |
|---|---|---|---|---|
| GPT-2 | hello, ! | வணக , ' ் , கம , ' ! | ආය , ' ු ', බ , ' ෝ , වන , '! | नमस , ' ् , त , ' े! |
| GPT-4 | hello, ! | வணக , ' ்கம , ' ! | ආය , ' ුබ , ' ෝවන , '! | नमस , ' ्त , ' े! |
| Llama 3 | hello, ! | வணக , ' ்கம , ' ! | ආය , ' ුබ , ' ෝවන , '! | नमस , ' ्त , ' े! |
| BERT | hello, ! | வணக்கம் , ! | ආයුබෝවන් , ! | नमस्ते , ! |
| MBERT | hello, ! | வணக்கம் , ! | ආයුබෝවන් , ! | नमस्ते , ! |
| T5 | hello! | வணக்கம்! | ආයුබෝවන්! | नमस्ते! |
| MT5 | hello! | வணக்கம்! | ආයුබෝවන්! | नमस्ते! |
| MBART | hello! | வணக்கம்! | ආයුබෝවන්! | नमस्ते! |
| NLLB | hello! | வணக்கம்! | ආයුබෝවන්! | नमस्ते! |

Table 1: This table displays the outputs of the pre-tokenizer from various language models for the same word in English, Tamil, Sinhala, and Hindi.

**Pre-tokenizers based on regular expressions.** GPT-2 popularized the use of large regular expressions to segment text into smaller chunks before applying BPE. Since then, many LLMs, including closed-source models like GPT-4 and open-weight models like Llama 3, have adopted pre-tokenization to segment text before training the tokenizer. This is especially common with BPE-based tokenizers.

**pre-tokenization as tokenization bounds.** The tokenization algorithm is applied to smaller pre-tokens rather than the entire text during the tokenizer training phase. As a result, the pre-tokenization step governs the maximum possible token length that can be learned from these pre-tokens. For example, consider the text "Hello World" and assume we use pre-tokenization by splitting on whitespaces. The pre-tokens will be "Hello" and "World". If we apply the BPE algorithm to these pretokens, the longest possible token learned from the first pre-token will be "Hello." However, due to data limitations, smaller tokens may be learned from within "Hello", such as "He", "l", and "lo" but the sequence cannot exceed "Hello". Thus, the pre-tokenizer effectively bounds the maximum token length that can be formed for a given pre-token. Consequently, the Compression Ratio ($CR$) calculated using Equation 1 and the Tokenization Parity ($TP$) calculated using Equation 2 become the Maximum Compression Ratio ($CR_{max}$) and the Minimum Tokenization Parity ($TP_{min}$), respectively.

### 3.2 Graphemes

Writing systems around the world vary in how they represent language, and they can be classified into six main types, namely logosyllabary/morphosyllabary, syllabary, abjad, alphabet, and abugida, based on the relationship between symbols and the spoken components of language (Daniels et al., 2003). Understanding the complexities involved in how these characters are represented in Unicode encoding is essential to processing these languages. For instance, in the abugida writing system, most of the characters are encoded using several Unicode points placed in a particular order, and when processing them, we need to treat those Unicode points together as a single unit, not separately. This sequence forms a character in the respective language, hereafter referred to as a grapheme. For instance, "ஶ்ரீ", a Grantha grapheme encoded in Tamil Unicode, is represented using four Unicode points corresponding to the following glyphs: 'ஶ', ' ் ', 'ர', and ' ீ '.[2] Similarly, the character ක්‍රෝ in the Sinhala language, which is also based on the abugida writing system, is a sequence of five Unicode points represented by the glyphs 'ක', ' ් ', 'x200D', 'ර ', and ' ෝ ', although it is considered a single character or grapheme in the language, where `'x200D'` is called Zero-Width Joiner. In these examples, glyphs like ' ී ', ' ් ', and ' ் ' are called vowel modifiers, which cannot stand alone and must always be processed along with the consonant to which they are attached. These are not diacritics but vowels, represented by vowel modifiers[3]. These modifiers take different shapes when joined with different consonants. The pulli – ' ் ' in Tamil – is not a standalone symbol but part of a pure consonant. For instance, 'க்' is the pure consonant 'k' in Tamil, and when a vowel (or vowel modifier), such as 'a', is added, the pulli will disappear. This understanding is important for NLP development, including tokenizers, because if an NLP tool breaks apart a vowel modifier, the result will not make much sense. Table 2 illustrates another example where how graphemes are formed in Abugida writing system based scripts.

---
[2] https://www.unicode.org/charts/PDF/U0B80.pdf

|              | **ta**        | **si**                | **hi**           |
|--------------|---------------|-----------------------|------------------|
| Unicode codepoint | ந , ன , ், ற, ி | ස , ්, ත , ු, ත,් , ව,් | ध , न , ्, य, व, ा, द |
| Grapheme Clusters | ந , ன் , றி    | ස්, තු , ති , ව       | ध , न् , य , वा    |

Table 2: The table demonstrates how characters are perceived when separated at the Unicode codepoint and the grapheme cluster level. This comparison is performed on the phrase "Thank you" translated into Tamil, Sinhala, and Hindi.

## 4 Software and Other Specifications.

**Software Specifications.** All code for the experiments was written in Python. We utilized tokenizers from the `tokenizers`[4] library, which is part of the HuggingFace (HF) Transformers framework (Wolf et al., 2020). For obtaining grapheme clusters, we used the `grapheme`[5] Python library.

**Tokenizers Used.** For our experiments, we use pretrained tokenizers from both English Centric (EC) and multilingual models. EC models include GPT-2 (Radford et al., 2019), GPT-4 (OpenAI et al., 2023), Llama 3 (Dubey et al., 2024), FLAN-T5 (Chung et al., 2022), and Gemma 2 (Team et al., 2024). Multilingual models include Aya (Üstün et al., 2024), multilingual BERT (Devlin et al., 2019) (referred to as mBERT), mT5 (Xue et al., 2021), mBART (Liu et al., 2020), and NLLB (Team et al., 2022). Note that while mBERT, mT5, and mBART have EC counterparts (BERT, T5, and BART, respectively), these EC models are excluded from our analysis as they yielded identical results to their multilingual versions. We also utilize Byte-level tokenizers, specifically ByT5 (Xue et al., 2022) and CANINE (Clark et al., 2022). All models are available on Hugging Face (HF), except for GPT-4. We obtained the pre-tokenization regular expression for GPT-4 from (Dagan et al., 2024). We specifically use tokenizers that include a pretokenizer, confirmed by checking `tokenizer.backend_tokenizer.pre_tokenizer` is not `None` in our implementation.

**Languages for Evaluation.** We primarily focus on three South Asian languages which are based on the Abugida writing system : Hindi (hi), Tamil (ta), and Sinhala (si).

**Training and Testing Data.** For training, we randomly sample 150k Tamil samples from the Samanantar Dataset (Ramesh et al., 2021). For testing, we use the FLORES+ (Team et al., 2022) development testsets for Tamil (ta), Hindi (hi), and Sinhala (si). We fix the vocabulary size to 5k.

## 5 Methodology

In this section, we have described two separate methodologies for 1) analyzing pre-tokenization and Byte-level tokenizers, and 2) our proposed Grapheme Pair Encoding (GPE) tokenization.

### 5.1 Methodology for Analyzing Pre-tokenization and Byte-Level Tokenizers

For this analysis, we rely on the metrics **Compression Ratio** (CR) and **Tokenization Parity** (TP). We define the Compression Ratio (CR) as:

$$\text{CR} = \frac{\text{Original Sequence Length}}{\text{Tokenized Sequence Length}} \quad (1)$$

---
[3] https://www.unicode.org/charts/PDF/U0B80.pdf
[4] https://github.com/huggingface/tokenizers
[5] https://github.com/alvinlindstam/grapheme

For Tokenization Parity (TP), we adopt the definition by Petrov et al. (2024). The parity of sentence $A$ ($s_A$) relative to sentence $B$ ($s_B$) by tokenizer $t$ is defined as:

$$\text{TP} = \frac{|t(s_A)|}{|t(s_B)|} \quad (2)$$

where $t(s_A)$ represents the tokenization of sentence $s_A$, and $|t(s_A)|$ denotes it's length. It could be stated that tokenizer $t$ achieves parity for $A$ with respect to $B$ when the *tokenizer parity* is close to 1 (Petrov et al., 2024).

Since we consider the pre-tokenization outputs for these calculations, as explained in Section 3.1, the calculated $CR$ represents the maximum $CR$ ($CR_{max}$), as pre-tokenization determines the maximum number of tokens present. By the same logic, the calculated Tokenization Parity ($TP$) represents the minimum $TP$ ($TP_{min}$).

**Analyzing Pretokenization.** We evaluate English Centric (EC) and Multilingual (ML) language models based on their pre-tokenization outputs rather than the final output of the tokenizers. Tokenizer training practices, including the choice of datasets and hyperparameters, vary across development teams. Consequently, comparisons of tokenizers based solely on their final outputs can be misleading. Instead, we focus on the pre-tokenization step, as detailed in Section 3.1. The longest sequence a tokenizer can learn in the pre-tokenization is a pretoken. Although smaller sub-words might be learned with insufficient training data, in an ideal case where ample data is provided, the learned tokens will match the pre-tokens. By analyzing the outputs of pre-tokenizers rather than those of the tokenizers themselves, we gain a clearer understanding of the ideal tokenization a tokenizer can achieve. This insight is crucial when assessing which tokenizers are best suited for complex script languages, particularly under the assumption that ample data will be provided when integrating a complex script language with the tokenizer. As shown in Figure 1, the impact of pre-tokenization is far more significant than that of the tokenization algorithm itself. Results of the pre-tokenization analysis are presented in Tables 3, 4, and 1.

**Analyzing Byte-level Tokenizers.** Since Byte-level tokenizers operate directly on the fundamental representation of text in digital form (bytes), analyzing them at the pre-tokenization level is not applicable. Therefore, we evaluate these tokenizers based on their final output. Results for Byte-level tokenizers are presented in Tables 5 and 6.

## 5.2 Grapheme Pair Encoding (GPE)

Instead of considering bytes as the smallest units, as done in BPE, we consider graphemes as the smallest units (refer Section 3.2). by introducing a preprocessing step that breaks the given text into graphemes and updates the initial vocabulary with the unique graphemes present in the tokenizer training data. Once the initial vocabulary is updated, the remainder of the method adheres to the standard BPE algorithm, but operates on graphemes. Our proposed methodology is detailed in Algorithm 1. We compare the GPE approach against vanilla implementations of BPE, Unigram, and WordPiece algorithms. All tokenizers are trained on a randomly sampled subset of 150k examples from the Samanantar Tamil dataset (Ramesh et al., 2021) and tested on the FLORES+ Tamil development testset (Team et al., 2022). For a fair comparison, we use a whitespace-based pre-tokenizer for all algorithms.

## 6 Results and Discussion

### 6.1 English Centric (EC) and Multilingual (ML) Models

Tables 3 and 4 present the maximum compression ratio ($CR_{\max}$) and the minimum tokenization parity ($TP_{\min}$), respectively (refer Section 3.1). The observations from both tables are discussed jointly, as they complement each other.

As expected, both English-centric (EC) and multilingual (ML) models demonstrate a strong compression ratio of 5× for the English language. This can be attributed to the simplicity of the English script; being part of the ASCII system, English characters fit into a single byte in UTF-8 representation.

The GPT-2 pre-tokenizer exhibits the lowest performance, with a $CR_{\max}$ score of only 1.6× across Tamil, Sinhala, and Hindi — all non-English languages. This indicates that the compression ratio achievable by a GPT-2 tokenizer, even with sufficient training data for these lan-

**Algorithm 1:** Grapheme Pair Encoding (GPE)

**Input:** Dataset $D$ of lines $N$,
Vocabulary size $|V|$,
pre-tokenization regular expression $RE$
**Output:** Vocabulary $V$, Merges $M$
// init Vocabulary and Merges
$V \leftarrow \{\}$
$M \leftarrow \{\}$
// init storing unique graphemes
$ghs \leftarrow \{\}$
**for** $i \leftarrow 1$ **to** $N$ **do**
    // extract pre-tokens based on RE
    $pre-tokens \leftarrow \text{Extract}(D[i], RE)$
    **for** *each $pt_i \in pretokens$* **do**
        // get graphemes $gs_i$ for each $pt_i$
        $gs_i \leftarrow \text{GetGraphemes}(pt_i)$
        // get unique graphemes from $gs_i$ compared to $ghs$
        $gs_{unique} \leftarrow \text{GetUnique}(gs_i, ghs)$
        // update $ghs$ with the new graphemes
        $ghs \leftarrow \text{Update}(gs_{unique}, ghs)$
    **end**
**end**
// update vocabulary with $ghs$
$V \leftarrow \text{Update}(ghs, V)$
// follow standard BPE merges
$V, M \leftarrow \text{BPE}(D, V, M)$

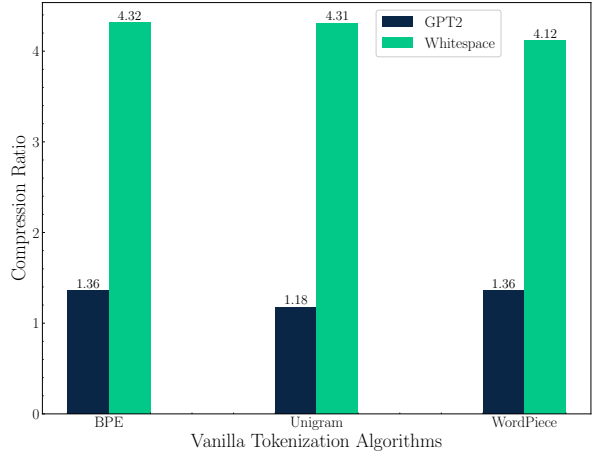

Figure 1: The figure illustrates the compression ratio of vanilla tokenizers trained on a randomly sampled 150k subset of the Samanantar Tamil dataset using GPT-2 and whitespace-based pre-tokenization, and tested on the FLORES+ Tamil development testset.

guages, will not exceed 1.6×. Similarly, GPT-2 demonstrates the highest $TP_{\min}$ across all compared non-English languages, with Tamil having a $TP_{\min}$ of 4.54. This result suggests that, on average, at least 4.54 Tamil tokens are required for every English token. The $TP$ can also be interpreted as a proxy for the relative context window size, implying that GPT-2 requires a context window 4.54 times longer for Tamil compared to English. Consequently, from a tokenization perspective, GPT-2 appears unsuitable for adapting to languages like Tamil, Sinhala, and Hindi. Both GPT-4 and Llama 3 show better performance than GPT-2, achieving a $CR_{\max}$ greater than 2× and a $TP_{\min}$ well below 3.0. However, despite these improvements, GPT-4 and Llama 3 still underperform when compared to the multilingual models, as well as the English-centric models FLAN-T5 and Gemma 2. We hope this observation serves as a cautionary note for researchers utilizing or adapting GPT-4 and Llama 3 for non-English languages.

As expected, multilingual models perform well on both the $CR_{\max}$ and $TP_{\min}$ metrics. Interestingly, FLAN-T5 and Gemma 2 perform on par with the best-performing multilingual models for non-English languages. This observation suggests that FLAN-T5 and Gemma 2 could be promising models for adapting to new languages. We leave the exploration of the impact of selecting non-English language-friendly pre-tokenization on language modeling for future work.

It is now evident that pre-tokenization plays a crucial role in how a language is represented within language models. To substantiate this claim, we trained vanilla tokenization algorithms — BPE, Unigram, and WordPiece—on a sufficiently large Tamil dataset (a randomly sampled 150k subset of the Samanantar dataset) with a vocabulary size of 5k. The training was conducted using either GPT-2 pre-tokenization or whitespace pre-tokenization, where the text is segmented based on whitespaces before being fed into the tokenization algorithm. Figure 1 illustrates the compression ratio of tokenizers trained under these condi-tions, tested using the FLORES+ development testset.

| Tokenizer |    | en   | ta   | si   | hi   |
|-----------|----|------|------|------|------|
| GPT-2     | EC | 5.26 | 1.36 | 1.55 | 1.56 |
| GPT-4     | EC | 5.23 | 2.13 | 2.16 | 2.04 |
| Llama 3   | EC | 5.23 | 2.13 | 2.16 | 2.04 |
| FLAN-T5   | EC | 6.06 | 9.21 | 6.34 | 5.13 |
| Gemma 2   | EC | 6.06 | 9.21 | 6.34 | 5.13 |
| Aya       | ML | 6.06 | 9.21 | 6.34 | 5.13 |
| mBERT     | ML | 5.22 | 7.77 | 5.63 | 4.59 |
| mT5       | ML | 6.06 | 9.21 | 6.34 | 5.13 |
| mBART     | ML | 6.06 | 9.21 | 6.34 | 5.13 |
| NLLB      | ML | 6.06 | 9.21 | 6.34 | 5.13 |

Table 3: The table shows the maximum compression ratio achievable for the four languages based on the pre-tokenization functions for both English-centric (EC) and multilingual (ML) models.

| Tokenizer |    | ta   | si   | hi   |
|-----------|----|------|------|------|
| GPT-2     | EC | 4.54 | 3.41 | 3.38 |
| GPT-4     | EC | 2.89 | 2.42 | 2.56 |
| Llama 3   | EC | 2.89 | 2.42 | 2.56 |
| FLAN-T5   | EC | 0.78 | 0.96 | 1.18 |
| Gemma 2   | EC | 0.78 | 0.96 | 1.18 |
| Aya       | ML | 0.78 | 0.96 | 1.18 |
| mBERT     | ML | 0.80 | 0.93 | 1.13 |
| mT5       | ML | 0.78 | 0.96 | 1.18 |
| mBART     | ML | 0.78 | 0.96 | 1.18 |
| NLLB      | ML | 0.78 | 0.96 | 1.18 |

Table 4: The table shows the minimum tokenization parity relative to English for the three languages in both English-centric (EC) and Multilingual (ML) models. This can also be interpreted as the minimum context window size relative to English.

The results of the tokenizers trained with GPT-2 pre-tokenization show nearly equal and consistently poor performance. Specifically, the BPE algorithm trained with GPT-2 pre-tokenization validates the predicted $CR_{\max}$ score in Table 3 for GPT-2; even with a sufficiently large dataset, the compression ratio did not exceed 1.36. In contrast, tokenizers trained using simple whitespace pre-tokenization outperform those trained with GPT-2 pre-tokenization by a significant margin. Moreover, all tokenizers trained with whitespace pre-tokenization exhibit similar performance levels.

Given that the tokenizers trained with GPT-2 and whitespace pre-tokenization methods show comparable performance within their respective groups, this finding demonstrates that the compression ratio is primarily determined by the pre-tokenization methodology employed, rather than the specific tokenization algorithm used.

### 6.2 Byte Level Tokenization

Tables 5 and 6 present a comparison of Byte-level tokenizers and Grapheme-based character extractor based on Compression Ratio ($CR$) and Tokenization Parity ($TP$). We utilize two byte-level tokenizers: ByT5, which employs UTF-8 encoding to break text into bytes, and CANINE, which uses UTF-32 encoding based on Unicode codepoints. This approach allows us to handle all Unicode characters, including rare symbols and emojis, without relying on extensive vocabularies or complex preprocessing. Comparing these byte-level methods with grapheme cluster-based character-level tokenization is fair and informative, as both tokenize text at a fundamental level. Our comparison highlights the advantages of aligning tokenization with graphemes, particularly for languages with complex scripts.

In Table 5, we observe that the $CR_{max}$ for CANINE is close to 1 for all languages, attributed to it's UTF-32 representation of each character. In contrast, ByT5, which uses UTF-8 encoding, performs poorly on non-English languages such as Tamil, Sinhala, and Hindi, as these scripts require multiple bytes per character in UTF-8 encoding scheme. Grapheme-based tokenization, however, shows a significant performance improvement, with $CR_{max}$ exceeding 1.4×. This improvement is due to graphemes effectively capturing characters by combining multiple codepoints into a single unit. Table 2 illustrates this with a comparison of how the phrase "Thank you" is represented at the Unicode codepoint and grapheme levels in Tamil, Sinhala, and Hindi.

Table 6 displays the minimum Tokenization Parity ($TP_{min}$) for the Byte-level and Grapheme cluster-based tokenizers. CANINE achieves near-parity with English for Tamil, Sinhala, and Hindi, indicating it requires a sim-ilar context window size for these languages. ByT5 performs the worst among the tokenizers, with Tamil having the lowest $TP_{min}$ of 3.2, meaning Tamil text requires three times the context window size of English. Sinhala and Hindi require at least 2.5×the context window size compared to English. In contrast, grapheme-based tokenization achieves the best performance, requiring less than 0.76 times the context window size of English for Tamil, Sinhala, and Hindi. These results suggest that grapheme-based tokenization is a superior choice for character-based tokenizers, especially for complex scripted languages. It enhances character representation by aligning more closely with human perceived characters. The impact of this choice on language modeling is a topic for future research and is beyond the scope of this paper.

| Tokenizer | en | ta | si | hi |
|---|---|---|---|---|
| CANINE | 0.98 | 0.99 | 0.98 | 0.98 |
| ByT5 | 0.99 | 0.37 | 0.38 | 0.39 |
| Grapheme based | 1.0 | 1.55 | 1.41 | 1.45 |

Table 5: Compression ratios for four languages using Byte-Level Tokenizers and our grapheme-based character extractor.

| Tokenizer | ta | si | hi |
|---|---|---|---|
| CANINE | 1.17 | 1.0 | 1.0 |
| ByT5 | 3.2 | 2.62 | 2.55 |
| Grapheme based | 0.76 | 0.71 | 0.69 |

Table 6: Tokenization parity relative to English for three languages using Byte-Level tokenizers and our grapheme-based character extractor.

### 6.3 Grapheme Pair Encoding (GPE)

We evaluate our proposed Grapheme Pair Encoding (GPE) alongside traditional tokenization algorithms such as BPE, Unigram, and WordPiece. To ensure a fair comparison, we utilize the Whitespace pretokenizer from HuggingFace. Details of the training process are outlined in Section 5.2.

| BPE | Unigram | WordPiece | GPE |
|---|---|---|---|
| 4.32 | 4.31 | 4.12 | **4.36** |

Table 7: Compression ratios of vanilla tokenization algorithms and GPE trained on a 150k Tamil sample from the Samanantar dataset and tested on the FLORES+ Tamil development testset.

Table 7 contains the Compression Ratio ($CR$) for Tamil from the aforementioned experiment. Our proposed method, GPE, better than all other tokenization algorithms, though the improvement is not significant. While GPE, being a derivative of the BPE algorithm, achieves a better $CR$, the difference of 0.04 is relatively minor. The WordPiece algorithm shows the poorest performance among them, but all algorithms achieve a $CR$ greater than 4. This supports the conclusion from Figure 1 that the pretokenizer has a more significant impact on Compression Ratio and Tokenizer Parity (and indirectly, context window size) than the choice of tokenization algorithm.

We have adapted the BPE algorithm to incorporate Graphemes, as the implementation is relatively straightforward. Future work will explore the integration of graphemes into other algorithms such as Unigram and WordPiece.

## 7 Conclusion

In this work, we demonstrate the crucial role of pre-tokenization in achieving an egalitarian representation of languages in language models. To the best of our knowledge, this is the first study to focus specifically on pretokenizers. Our findings reveal that popular English-Centric language models inadequately represent complex scripted languages like Tamil, Sinhala, and Hindi. This underscores the need for caution when using English-Centric models as the base for developing language-specific LLMs. We advocate for the exploration of tokenization choices used by multilingual models, as they are progressing towards more egalitarian language representation.

Our analysis shows that pre-tokenization significantly affects tokenization, often more than the choice of algorithm. We propose improving BPE by incorporating graphemes, creating our Grapheme Pair Encoding (GPE) method. However, our focus is on the tokenizer, leaving broader implications for language model performance to future research. We hope this work sparks further exploration and advances the goal of egalitarian language representation in language modeling. The code bases will be provided upon request.